%% file: main.tex
\documentclass[letterpaper, 10 pt, conference]{ieeeconf}  

\IEEEoverridecommandlockouts                     
\usepackage{graphics} 
\usepackage{epsfig} 
\usepackage{mathptmx} 
\usepackage[utf8]{inputenc}
\usepackage[T1]{fontenc}
\usepackage{booktabs}
\usepackage{amsfonts}
\usepackage{nicefrac}
\usepackage{microtype}
\usepackage{times} 
\usepackage{amsmath} 
\usepackage{amssymb}  
\usepackage{xspace}
\usepackage{booktabs}
\usepackage{multirow}
\usepackage{threeparttable}
\usepackage{afterpage}
\usepackage{caption}
\usepackage{cuted}
\usepackage{graphicx}
\usepackage{multirow}
\usepackage{wrapfig}
\usepackage{subcaption}
\usepackage{algorithm}
\usepackage{algorithmic}

\newcommand{\eg}{\emph{e.g.}\xspace}

\newcommand{\R}{\mathbb{R}}
\newcommand{\method}{GST-VLA\xspace}

\usepackage{cite}
\makeatletter
\let\NAT@parse\undefined
\makeatother
\usepackage[colorlinks=true,   
            linkcolor=blue,   
            citecolor=blue,   
            urlcolor=blue,    
            pdfborder={0 0 0} 
]{hyperref}

\title{\LARGE \bf
GST-VLA: Structured Gaussian Spatial Tokens for 3D Depth-Aware Vision-Language-Action Models
}

\author{%
    Md Selim Sarowar$^{1}$, 
    Omer Tariq$^{2}$,
    Sungho Kim*$^{1}$
    \\[2ex]
    $^{1}$Yeungnam University\\
    $^{2}$Korea Advanced Institute of Science and Technology, KAIST
    \\
}

\begin{document}

\maketitle
\thispagestyle{empty}
\pagestyle{empty}


\begin{strip}
    \centering
    \includegraphics[width=1.0\textwidth]{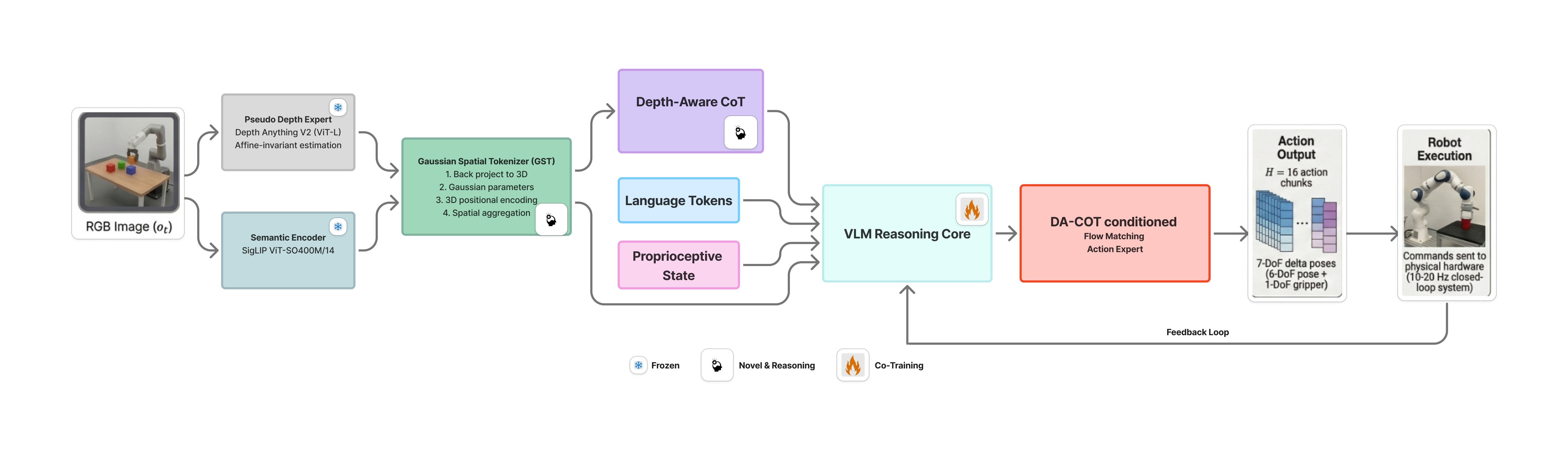}
    \captionof{figure}{The proposed GST-VLA pipeline integrates five sequential stages to ground robot actions in structured
  3D spatial reasoning. A frozen semantic encoder and a frozen depth expert process the RGB observation in parallel,
  extracting dense patch features and affine-invariant metric depth respectively. The novel trainable Gaussian Spatial
  Tokenizer (GST) fuses these streams by back-projecting depth into 3D, estimating per-patch Gaussian parameters
  $(\mu, \sigma, \alpha)$ from visual features, applying 3D Fourier positional encoding, and aggregating to
  $N_g$ structured spatial tokens via spatial attention pooling. These tokens are projected into the VLM reasoning
  core through a cross-attention projector, which generates supervised Depth-Aware Chain-of-Thought (DA-CoT)
  intermediate reasoning over 3D object grounding, grasp affordance, metric spatial relations, and SE(3) motion
  plan waypoints before producing action conditioning tokens.}
    \label{fig:main_teaser}
\end{strip}

\begin{abstract}
VLA models encode visual observations as 2D patch tokens with no intrinsic geometric structure. Augmenting with dense monocular depth, as in DepthVLA, injects pixel-uniform scalar values that encode neither surface orientation nor geometric confidence, and provides no mechanism for intermediate spatial verification before action decoding. We introduce \method with two contributions. First, the Gaussian Spatial Tokenizer (GST) converts frozen dense depth and frozen semantic patch features into $N_g{=}128$ anisotropic 3D Gaussian primitives, each parameterized by a metric residual mean $\mu \in \R^3$, log-scale covariance $\sigma \in \R^3$, and learned opacity
$\alpha \in (0,1)$. The covariance eigenstructure encodes local surface orientation, and opacity provides per-primitive geometric confidence, both inaccessible from scalar depth. Spatial attention pooling with learned queries concentrates the fixed token budget on geometrically salient regions
rather than distributing uniformly. Second, Depth-Aware Chain-of-Thought (DA-CoT) reasoning supervises four structured intermediate spatial thoughts, covering 3D object grounding, grasp affordance contact geometry, pairwise metric distances, and coarse SE(3) waypoints, as explicit generation targets in the training loss. A cross-attention sublayer at every VLM transformer block provides direct access to the raw 256-primitive Gaussian field during DA-CoT generation. A 300M-parameter flow-matching action expert with mixture-of-experts feedforward sublayers decodes 7-DoF delta action chunks via conditional ODE integration, conditioned on both VLM hidden states and DA-CoT outputs through dual cross-attention. Trained with composite
$\mathcal{L}_\mathrm{flow} + \mathcal{L}_\mathrm{CoT} + \mathcal{L}_\mathrm{depth}$
across three progressive stages, \method achieves 96.4\% on LIBERO (+2.0\%), and 80.2\% on SimplerEnv (+5.4\%). Ablations isolate the contribution of each GST component, each DA-CoT thought, and each training stage, confirming independent and synergistic gains concentrated on precision-demanding tasks.

\end{abstract}

\input{introduction}

\input{related_work}

\input{method}

\input{experiments}

\section{Conclusion}
\label{sec:conclusion}

\method introduces two contributions for 3D-grounded VLA models. The Gaussian Spatial Tokenizer converts frozen depth and visual features into anisotropic 3D Gaussian tokens whose covariance eigenstructure encodes surface orientation and whose learned opacity encodes geometric confidence, with spatial attention pooling concentrating representational capacity on task-relevant geometry. 
Depth-Aware Chain-of-Thought reasoning supervises explicit 3D spatial verbalization as an intermediate generation target, with cross-attention to the raw Gaussian field providing
geometric access at full resolution during each thought. The composite training loss couples the CoT and depth objectives with the Gaussian field parameters, creating synergistic gradient pathways between reasoning quality and geometric calibration. Evaluations and ablations confirm that each component contributes independently and that their combination provides super additive gains concentrated on precision demanding manipulation tasks.

\bibliographystyle{IEEEtran}
\bibliography{references}
\section*{APPENDIX}
The results presented in this paper are preliminary. Please note that the experiments are currently ongoing, and the final data is subject to change upon the completion of the study.
All ideas, results, methods, and any content herein are the sole property of the authors. Reuse, reproduction, distribution, or any other use without explicit written permission from the authors is strictly prohibited. All rights reserved.

\section*{ACKNOWLEDGMENT}

This work was supported by Regional Innovation System \& Education(RISE) program[B0080529002330], through the Gyeongbuk RISE CENTER, funded by the Ministry of Education(MOE) and the Gyeongsangbuk-do, Republic of Korea.(2025-RISE-15-115)


\end{document}

%% file: introduction.tex
\section{INTRODUCTION}

VLA models~\cite{kim2024openvla,zhao2025cotvla,qu2025spatialvla,yuan2025depthvla,zhen20243dvla} fine-tune large VLMs on demonstration data to produce robot control policies. Visual observations enter these models as 2D patch tokens: a grid of $N_p$ fixed-resolution embeddings capturing local
appearance statistics. Each patch token occupies a fixed spatial extent in pixel space irrespective of the underlying scene geometry. No patch token encodes depth, surface normal direction, or geometric confidence. When manipulation requires millimeter scale geometric accuracy, such as edge grasping, peg insertion, or thin object picking, the model must recover 3D structure implicitly within its hidden states, a computation that degrades systematically as task precision
increases.

DepthVLA~\cite{yuan2025depthvla} addresses this by adding a depth expert as a third transformer
stream in a mixture-of-transformers (MoT) design. The depth stream shares attention layers with
the VLM and action expert, enabling action tokens to attend to intermediate depth representations
rather than a final depth map. This establishes that explicit geometric signals improve
manipulation precision. However, three structural limitations remain. (i) The depth
representation is pixel uniform: each token holds a scalar depth value at a fixed pixel location,
distributing token budget equally across geometrically relevant and irrelevant regions. (ii) No token encodes surface orientation. Scalar depth at a point provides no information about the local tangent plane; a flat surface and a sharp edge at identical depth produce identical representations. (iii) There is no mechanism for the model to explicitly verify or articulate its 3D scene interpretation before generating actions. The spatial reasoning pathway from depth tokens to action tokens is fully implicit and non-inspectable.

\method addresses all three limitations. The Gaussian Spatial Tokenizer (GST) replaces the dense
scalar depth stream with $N_g{=}128$ anisotropic 3D Gaussian primitives. Each primitive
$(\mathbf{c}_k, \Sigma_k, \alpha_k)$ is a volumetric spatial token whose axis-aligned covariance
$\Sigma_k = \mathrm{diag}(\exp(2\sigma_k))$ encodes surface orientation through its eigenstructure
(the minimum-eigenvalue axis approximates the surface normal), whose opacity $\alpha_k \in (0,1)$
encodes geometric confidence (suppressing tokens on specular or textureless surfaces where the
depth estimate is unreliable), and whose spatial allocation via learned attention pooling
concentrates the fixed token budget on task-relevant geometry. Depth-Aware Chain-of-Thought
(DA-CoT) reasoning introduces a supervised intermediate generation stage: the VLM must produce
four structured spatial thoughts (3D object centroid, grasp contact geometry, metric spatial
relations, SE(3) waypoints) as explicit supervised targets before generating action-conditioning
tokens. During this generation, a cross-attention sublayer at every VLM transformer block provides
unfiltered access to the full 256-primitive raw Gaussian field, allowing thought-level queries into
specific geometric regions.

Our contributions are summarized as follows:

\begin{itemize}

\item \textbf{GST-VLA architecture: }The GST, a trainable module producing structured 3D Gaussian tokens from frozen depth and visual features via back-projection, per-patch parameter estimation with multi-scale opacity gating, 3D Fourier positional encoding, and spatial attention pooling.A three-stage training protocol with composite
$\mathcal{L}_\mathrm{flow} + \mathcal{L}_\mathrm{CoT} + \mathcal{L}_\mathrm{depth}$ objective

\item \textbf{DA-COT:} DA-CoT, a supervised intermediate reasoning stage imposing explicit 3D geometric targets as
sequential generation within the VLM.

\item \textbf{Data Efficient validation:} We demonstrate that GST-VLA with lower computation cost \& parameters, significantly outperforms state-of-the-art VLAs in simulated environments (LIBERO, Simpler), achieving notable gains in grasping accuracy, collision avoidance, and overall task success.

\end{itemize}

%% file: related_work.tex
\section{Related Work}
\subsection{Spatial Representations in VLA Models}

DepthVLA~\cite{yuan2025depthvla} introduces a depth expert as a dedicated transformer stream within a MoT architecture, pretrained on metric depth datasets and jointly trained with the VLM and action expert. The action expert attends to intermediate depth features at every transformer layer, not merely a final depth map. SpatialVLA~\cite{qu2025spatialvla} injects 3D egocentric position encodings derived from an off-the-shelf depth estimator into patch tokens, modulating the 2D
features with scalar position offsets, and introduces adaptive spatial action grids for discrete action tokenization. Both approaches share a structural property: depth enters the model as a pixel-uniform scalar, with one value per spatial location. The GST departs from this by parameterizing each spatial token as a full anisotropic Gaussian primitive with seven learned
parameters (three for mean offset, three for log-scale covariance, one for opacity), and compressing $N_p{=}256$ raw tokens into $N_g{=}128$ via learned spatial attention pooling rather than uniform spatial binning. This produces tokens that encode surface orientation and geometric confidence, and
that concentrate representational capacity on task relevant regions.

\subsection{3D Gaussian Primitives Beyond Rendering}

3D Gaussian Splatting~\cite{kerbl20233dgs} represents scenes as sets of anisotropic Gaussian primitives $\{(\mu_k, \Sigma_k, \alpha_k)\}$ optimized via differentiable rendering for novel view synthesis. GaussTR~\cite{jiang2025gausstr} aligns Gaussian representations with foundation model features for self-supervised 3D understanding. GPSToken~\cite{zhang2025gpstokengaussianparameterizedspatiallyadaptive} uses 2D Gaussian functions for spatially-adaptive image tokenization, decoupling spatial layout from texture for generation. The GST departs from all of these in objective: it does not optimize for rendering, understanding, or image tokenization, but uses the Gaussian primitive as a spatial token format consumed by a VLM for manipulation policy conditioning. The Gaussian parameterization is valued not for its rendering properties but for the geometric information it encodes per token: position,
anisotropic extent, and confidence, all calibrated against metric depth through a differentiable rendering auxiliary loss.

\subsection{Chain-of-Thought for Embodied Reasoning}

Chain-of-thought reasoning~\cite{zhao2025cotvla} supervises intermediate steps in LLMs. ECoT~\cite{Zawalski24-ecot} studies spatial reasoning chains in manipulation settings. CogACT~\cite{li2024cogact} decouples cognition from action by conditioning a diffusion transformer on a cognition token output by the VLM. HybridVLA~\cite{liu2025hybridvla} unifies autoregressive and diffusion action generation within a single LLM but does not introduce structured spatial reasoning targets. None of these works supervise intermediate generation with explicit metric 3D coordinates, grasp contact geometry, or SE(3) waypoints derived from demonstration data. DA-CoT
introduces exactly this: four structured thought components, each supervised against offline 3D annotations, each generating tokens that flow through the VLM's autoregressive decoding before
action-conditioning tokens are produced.

%% file: method.tex
\section{Method}
\label{sec:method}

\method processes an RGB observation $o_t \in \R^{H \times W \times 3}$ ($H{=}W{=}224$), a
language instruction $\ell$, and proprioceptive state $s_t \in \R^7$ through five sequential
stages. A frozen visual encoder produces dense semantic patch features $\mathbf{F}_\mathrm{sem} \in \R^{256 \times 1152}$. A frozen monocular depth estimator produces affine invariant metric depth $\hat{D} \in \R^{H \times W}$. Both encoders remain frozen throughout all training stages. The two streams are fused exclusively within the GST.

\subsection{Gaussian Spatial Tokenizer}
\label{sec:gst}

The GST is a trainable module that converts the pair
$(\mathbf{F}_\mathrm{sem}, \hat{D})$ into $N_g$ structured volumetric spatial tokens. Four sequential operations construct the representation.

\subsubsection{Depth Back-Projection to Metric 3D Anchors}

Given metric depth $\hat{D}$ and calibrated camera intrinsics $K \in \R^{3 \times 3}$, each pixel
$(u, v)$ is lifted to camera frame 3D via
\begin{equation}
  \mathbf{p}_{uv} = \hat{D}_{uv} \cdot K^{-1} [u,\; v,\; 1]^\top \in \R^3.
  \label{eq:backproject}
\end{equation}
For each of the $N_p = 256$ semantic patches (corresponding to a $16{\times}16$ pixel receptive field at input resolution $224{\times}224$), we compute the mean of back-projected 3D coordinates within the receptive field, yielding metric anchors $\{\mathbf{p}_k\}_{k=1}^{N_p}$. Each anchor
$\mathbf{p}_k$ localizes patch $k$ in the camera-frame 3D coordinate system. The anchor precision is bounded by two factors: the depth estimator accuracy (median relative error $\sim$3\% on indoor scenes) and the spatial averaging within the $16{\times}16$ receptive field, which smooths
sub-patch depth variation. The residual offset $\mu_k$ estimated in the next step compensates for this averaging artifact.

\subsubsection{Per-Patch Gaussian Parameter Estimation}

A 4-layer MLP $f_\theta$ with hidden dimensions $[1152, 768, 512, 7]$ and GELU activations maps each semantic patch embedding to a 7-dimensional Gaussian parameterization:
\begin{equation}
  [\mu_k,\; \sigma_k,\; \hat{\alpha}_k] = f_\theta(\mathbf{F}_{\mathrm{sem},k}),
\end{equation}
decomposed as follows.

The residual mean $\mu_k \in \R^3$ is an offset from the back-projected anchor, yielding the primitive centroid $\mathbf{c}_k = \mathbf{p}_k + \mu_k$. This residual formulation is
structurally important: it decouples the coarse metric localization (provided by the depth estimator through $\mathbf{p}_k$) from fine geometric refinement (learned by the MLP through $\mu_k$). Ablating $\mu_k$ by fixing it to zero pins centroids to back-projected anchors and costs 1.9 percentage points (Table~\ref{tab:ablation_gst}), confirming that sub-patch refinement contributes meaningfully to downstream task performance.

The log-scale $\sigma_k \in \R^3$ parameterizes an axis-aligned anisotropic covariance $\Sigma_k = \mathrm{diag}(\exp(2\sigma_k))$. The three eigenvalues of $\Sigma_k$ encode the spatial extent of the primitive along each camera-frame axis. For primitives on flat surfaces, the eigenvalue corresponding to the surface normal direction is small (tight extent perpendicular to the surface), while the two tangential eigenvalues are large (diffuse extent along the surface).
For primitives on edges or corners, multiple eigenvalues contract. This geometric information is absent from scalar depth representations: two surface regions at identical depth but with different local curvature produce identical depth tokens in DepthVLA but distinct covariance structures in
the GST. Replacing anisotropic $\Sigma_k$ with isotropic covariance ($\sigma_k^x = \sigma_k^y = \sigma_k^z$) costs 1.6 percentage points (Table~\ref{tab:ablation_rep}), quantifying the contribution of orientation encoding.

The pre-activation opacity logit $\hat{\alpha}_k \in \R$ is processed through a multi-scale pathway. Rather than computing opacity from the patch feature alone, we aggregate across three
spatial scales via a multi-scale image pyramid (MIP):
\begin{equation}
  \alpha_k = \sigma\!\left(f_\mathrm{exp}\!\left(\mathrm{MIP}(\mathbf{F}_{\mathrm{sem},k})\right)\right) \in (0, 1),
  \label{eq:opacity}
\end{equation}
where $\mathrm{MIP}(\cdot)$ concatenates the patch feature with its $2{\times}$ and $4{\times}$ average-pooled spatial neighborhoods, producing a $3 \times 1152 = 3456$-dimensional input to a 2-layer opacity MLP $f_\mathrm{exp}: \R^{3456} \to \R^1$. The multi-scale context is necessary because geometric confidence at a patch location depends on surrounding texture gradient magnitude: a patch on a uniform surface surrounded by other uniform patches (low texture gradient at all scales) should receive low opacity because the depth estimate there lacks photometric verification. A patch on a textured object surrounded by background (high gradient locally, low
globally) should receive high opacity. The MIP aggregation provides exactly this scale-dependent context. Ablating opacity by fixing $\alpha_k = 1$ costs 1.5 percentage points
(Table~\ref{tab:ablation_gst}), confirming the value of confidence based token weighting.

\subsubsection{3D Fourier Positional Encoding}

The metric centroid $\mathbf{c}_k$ is encoded via multi octave 3D sinusoidal features:
\begin{equation}
  \mathrm{PPE}(\mathbf{c}_k) = \bigl[\sin(2^l \pi \mathbf{c}_k),\; \cos(2^l \pi \mathbf{c}_k)\bigr]_{l=0}^{L-1} \in \R^{6L},
  \label{eq:ppe}
\end{equation}
with $L = 6$ octaves yielding a 36-dimensional positional code. The choice of 3D Fourier encoding over learned 2D positional embeddings is motivated by a specific requirement: the VLM must be able to compute approximate metric distances between tokens by operating on their positional codes. Two tokens at centroids $\mathbf{c}_i$ and $\mathbf{c}_j$ separated by distance $d$ produce Fourier features whose inner product structure encodes $d$ across multiple frequency bands. Learned 2D positional embeddings encode pixel-space proximity, which conflates depth variation with lateral displacement: two objects at the same pixel column but different depths receive similar 2D
positions but require distinct 3D treatment. The 3D Fourier encoding resolves this conflation. Replacing 3D Fourier PE with learned 2D PE costs 2.8 percentage points (Table~\ref{tab:ablation_gst}), the largest single-component ablation within the GST.

The per-primitive spatial token is formed by projecting the concatenation
$[\mathbf{F}_{\mathrm{sem},k};\; \mathrm{PPE}(\mathbf{c}_k);\; \sigma_k;\; \alpha_k]
\in \R^{1192}$ (where $1192 = 1152 + 36 + 3 + 1$) through a learned linear projection
$\mathbf{W}_\mathrm{tok} \in \R^{1192 \times d_g}$ with $d_g = 768$ to produce the raw spatial
token for patch $k$.

\subsubsection{Spatial Attention Pooling}

The $N_p = 256$ raw spatial tokens are compressed to $N_g = 128$ output tokens via single-layer cross-attention:
\begin{equation}
  \mathbf{Z}_\mathrm{spatial} = \mathrm{softmax}\!\left(\frac{\mathbf{Q}_\mathrm{pool} \mathbf{K}^\top}{\sqrt{d_g}}\right) \mathbf{V} \in \R^{N_g \times d_g},
  \label{eq:attnpool}
\end{equation}
where $\mathbf{Q}_\mathrm{pool} \in \R^{N_g \times d_g}$ are $N_g$ learned pooling queries, $\mathbf{K}, \mathbf{V} \in \R^{N_p \times d_g}$ are key and value projections of the raw token set $\mathbf{T}_\mathrm{raw}$. Each learned query specializes to attend to a specific geometric
pattern in the raw token set. Queries that correspond to object surfaces attend to dense clusters of similar depth, high opacity, anisotropic primitives. Queries that correspond to background attend to scattered, high-$\sigma$, low-$\alpha$ primitives. The resulting allocation is adaptive: a scene with many small objects distributes pooling queries across multiple object surfaces; a scene with one large object and empty background concentrates queries on the object
and assigns few to background. This contrasts with the fixed uniform allocation of pixel-aligned depth, where token budget is spent equally on the task-relevant cup handle and the far wall. Replacing spatial attention pooling with uniform average pooling costs 2.1 percentage points
(Table~\ref{tab:ablation_gst}).

The full raw token set $\mathbf{T}_\mathrm{raw} \in \R^{N_p \times d_g}$ is retained separately
and used as the geometric reference for DA-CoT cross-attention
(Section~\ref{sec:dacot}).

\subsubsection{Differentiable Depth Rendering as Geometric Regularizer}

The predicted Gaussian field must remain geometrically consistent with the metric depth target. We enforce this through a differentiable rendering loss. For camera ray $\mathbf{r}$ with direction
$\mathbf{d}$ and origin $\mathbf{o}$, rendered depth is
\begin{equation}
  \hat{D}_\mathrm{render}(\mathbf{r}) = \sum_{k=1}^{N_p} w_k \cdot (\mathbf{c}_k^\top \mathbf{d}),
  \label{eq:render}
\end{equation}
where $w_k = \alpha_k^{(\mathbf{r})} \prod_{j < k}(1 - \alpha_j^{(\mathbf{r})})$ and the
ray specific opacity is
$\alpha_k^{(\mathbf{r})} = \alpha_k \exp\!\left(-\tfrac{1}{2}
\tfrac{(t_k - \mathbf{c}_k^\top \mathbf{d})^2}
{\mathbf{d}^\top \Sigma_k \mathbf{d}}\right)$
with $t_k = \|\mathbf{c}_k - \mathbf{o}\|$. Equation~\eqref{eq:render} is fully differentiable
with respect to $\mu_k$, $\sigma_k$, and $\alpha_k$, meaning that the depth reconstruction loss
$\mathcal{L}_\mathrm{depth}$ provides gradients that geometrically calibrate all GST parameters. This loss acts as a geometric regularizer: without it, the MLP could produce Gaussian parameters that are useful for downstream action prediction (via $\mathcal{L}_\mathrm{flow}$) but
geometrically inconsistent with the actual scene, \eg placing centroids at incorrect depths while compensating through covariance. The depth rendering loss prevents this degenerate solution.

\subsection{VLM Reasoning Core and DA-CoT}
\label{sec:dacot}

\subsubsection{Spatial Token Injection}

The pooled spatial tokens $\mathbf{Z}_\mathrm{spatial} \in \R^{N_g \times d_g}$ are projected into the VLM hidden dimension $d_\mathrm{VLM}$ via a two-layer cross-attention projector $\Psi$ with residual connection:
\begin{equation}
  \tilde{\mathbf{Z}} = \Psi(\mathbf{Z}_\mathrm{spatial}) =
  \mathrm{CA}\!\left(\mathbf{Z}_\mathrm{spatial},\; \mathbf{F}_\mathrm{sem}\right)\,
  \mathbf{W}_\mathrm{proj},
  \label{eq:proj}
\end{equation}
where $\mathbf{W}_\mathrm{proj} \in \R^{d_g \times d_\mathrm{VLM}}$. The cross-attention in Eq.~\eqref{eq:proj} re-grounds each spatial token in the full semantic feature space before VLM ingestion: a spatial token representing a cup handle re-attends to the semantic patch features that encode ``cup'' appearance, fusing geometric localization with object identity. The VLM input sequence is $\mathcal{X} = [\tilde{\mathbf{Z}};\; \mathbf{L}_\mathrm{tokens};\; \mathbf{e}_s]$
(spatial tokens, language embeddings, proprioceptive embedding). The VLM is adapted via LoRA
($r{=}16$, $\alpha{=}32$) on all self-attention projection matrices.

\subsubsection{Depth-Aware Chain-of-Thought}
\label{sec:dacot_detail}

Standard VLAs collapse 3D scene interpretation and action generation into a single undifferentiated computation within the VLM hidden states. There is no explicit representation of the model's 3D understanding that can be supervised, inspected, or verified. DA-CoT introduces a supervised intermediate generation stage that separates these two computations.

The separation is implemented architecturally. During DA-CoT token generation, an additional cross-attention sublayer is inserted at every VLM transformer block:
\begin{equation}
  \mathbf{h}_i^{(\ell)} \leftarrow \mathbf{h}_i^{(\ell)} +
  \mathrm{CA}\!\left(\mathbf{h}_i^{(\ell)},\; \mathbf{T}_\mathrm{raw},\; \mathbf{T}_\mathrm{raw}\right),
  \label{eq:dacot_ca}
\end{equation}
where $\mathbf{h}_i^{(\ell)}$ is the hidden state of token $i$ at VLM layer $\ell$, and $\mathbf{T}_\mathrm{raw} \in \R^{N_p \times d_g}$ is the full raw Gaussian token set (not the
pooled version $\mathbf{Z}_\mathrm{spatial}$). The use of raw rather than pooled tokens in Eq.~\eqref{eq:dacot_ca} is a deliberate design choice: spatial attention pooling compresses
$N_p{=}256$ tokens into $N_g{=}128$ for efficient VLM processing, but during CoT generation the model needs to query specific geometric regions at full resolution. For instance, generating the grasp contact point $c_2$ requires attending to the specific subset of primitives covering the object's graspable surface, which may correspond to only 3-5 raw tokens whose information is diluted by pooling. The cross-attention in Eq.~\eqref{eq:dacot_ca} provides this targeted geometric access.

The DA-CoT output is a structured chain $\mathcal{C} = (c_1, c_2, c_3, c_4)$ generated autoregressively before action-conditioning tokens. The four components are ordered by causal
dependency: each subsequent thought depends on information produced by preceding thoughts.

$c_1$: 3D object grounding. The model generates the metric centroid of the task relevant object
in camera coordinates, \eg ``target centroid: (0.15, $-$0.08, 0.42) m.'' This requires the VLM to
fuse the language instruction (identifying which object) with the Gaussian field (localizing it in
3D). The centroid estimate anchors all subsequent spatial reasoning; errors here propagate through
$c_2$, $c_3$, and $c_4$. Ablating $c_1$ costs 1.9 percentage points (Table~\ref{tab:ablation_cot}).

$c_2$: Grasp affordance. The model generates a 3D contact point offset relative to the $c_1$ centroid and the approach surface normal direction. This specifies where and at what angle the
gripper should engage the target. The contact point requires attending to the specific primitives on the object's graspable surface, querying their covariance eigenstructure to infer local surface orientation for the approach vector.

$c_3$: Metric spatial relations. The model generates signed metric distances between task-relevant objects and surfaces in camera frame. For a "place cup on shelf" task, $c_3$ might produce the vertical distance from the cup centroid to the shelf surface and the lateral distance to the shelf edge. These distances condition the action expert's trajectory height and lateral offset.

$c_4$: SE(3) motion plan. The model generates a coarse sequence of 6-DoF end-effector waypoints (pre-grasp, grasp, post-grasp retract) as $[\Delta x, \Delta y, \Delta z, \Delta r_x, \Delta r_y, \Delta r_z]$ deltas in camera frame. These waypoints provide the action expert with a geometric prior over trajectory shape. The action expert refines this coarse plan through flow matching, but the prior substantially constrains the search space. Ablating $c_4$ has the largest individual effect among the four components at $-2.3$ percentage points (Table~\ref{tab:ablation_cot}).

The full VLM output is $(\mathcal{C}, \mathbf{L}_\mathrm{action})$, where
$\mathbf{L}_\mathrm{action} \in \R^{N_a \times d_\mathrm{VLM}}$ are action-conditioning tokens generated after the CoT chain. The sequential ordering ensures that action tokens are generated in a context that includes the explicit 3D reasoning outputs, allowing the VLM to condition action generation on verified spatial understanding.

\begin{figure*}[t]
  \centering
  \includegraphics[width=1\textwidth]{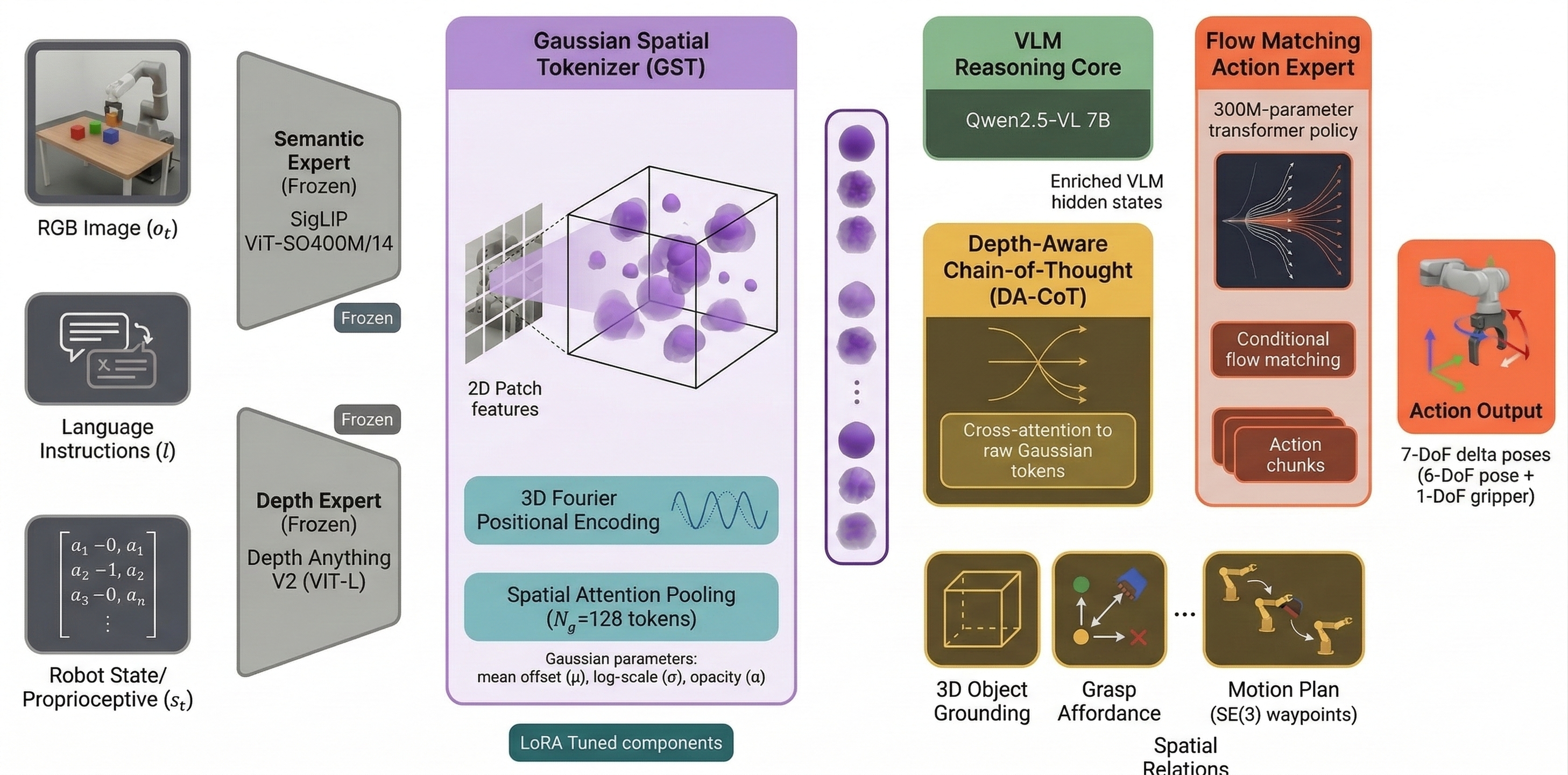}
  \caption{\method framework. Two frozen encoders produce semantic features and metric depth. The LoRA adapted GST lifts these into $N_g{=}128$ anisotropic 3D Gaussian tokens via four operations. A cross-attention projector injects spatial tokens into the VLM, where DA-CoT sublayers attend to the raw 256-primitive Gaussian field. The action expert receives dual conditioning: VLM hidden states (semantic and visual context) and DA-CoT action tokens (3D geometric reasoning).}
  \label{fig:architecture}
\end{figure*}

\subsection{Flow-Matching Action Expert}
\label{sec:policy}

The action expert is a 300M-parameter transformer with 6 layers at hidden dimension $d_e = 512$. Each layer contains a self-attention block, two cross-attention blocks (one attending to VLM hidden states $\mathbf{H}_\mathrm{vlm}$, one attending to DA-CoT action tokens $\mathbf{L}_\mathrm{action}$), and a mixture-of-experts (MoE) feedforward block with 8 experts per layer, top-2 routing via a learned gating network $g: \R^{512} \to \R^8$, and expert hidden dimension 2048 with SiLU activation. The dual cross-attention conditioning is:
\begin{equation}
  \mathbf{H}_\mathrm{act}^{(\ell)} = \mathbf{H}_\mathrm{act}^{(\ell)} +
    \mathrm{CA}\!\left(\mathbf{H}_\mathrm{act}^{(\ell)}, \mathbf{H}_\mathrm{vlm}\right)
  + \mathrm{CA}\!\left(\mathbf{H}_\mathrm{act}^{(\ell)}, \mathbf{L}_\mathrm{action}\right),
  \label{eq:expert_cond}
\end{equation}
where $\mathbf{H}_\mathrm{act}^{(0)} = \mathbf{e}_s$ (proprioceptive embedding). The two conditioning streams are functionally distinct: $\mathbf{H}_\mathrm{vlm}$ carries the VLM's fused semantic-visual representation; $\mathbf{L}_\mathrm{action}$ carries the 3D geometric reasoning
distilled by DA-CoT. Removing the $\mathbf{L}_\mathrm{action}$ stream costs 3.1 percentage points (Table~\ref{tab:ablation_stages}), confirming that DA-CoT outputs encode geometric information that is not redundant with what $\mathbf{H}_\mathrm{vlm}$ already captures.

The action distribution is modeled via conditional flow matching~\cite{lipman2022flow}. Given straight-line interpolation $\mathbf{a}_t = (1{-}t)\mathbf{a}_0 + t\mathbf{a}_1$ with $\mathbf{a}_0 \sim \mathcal{N}(\mathbf{0}, \mathbf{I})$, the velocity field $v_\theta: \R^{7 L_\mathrm{act}} \times [0,1] \to \R^{7 L_\mathrm{act}}$ is trained by:
\begin{equation}
  \mathcal{L}_\mathrm{flow} = \mathbb{E}_{t, \mathbf{a}_0, \mathbf{a}_1}
  \bigl[\|v_\theta(\mathbf{a}_t, t \mid \mathbf{H}_\mathrm{act}) - (\mathbf{a}_1 - \mathbf{a}_0)\|^2\bigr].
  \label{eq:flow}
\end{equation}
At inference, the ODE $\nicefrac{d\mathbf{a}_t}{dt} = v_\theta(\mathbf{a}_t, t \mid \mathbf{H}_\mathrm{act})$ is
integrated from $t{=}0$ to $t{=}1$ with 10 Euler steps. The model predicts action chunks of $L_\mathrm{act} = 10$ future 7-DoF delta poses (6-DoF end effector plus 1-DoF gripper), with
temporal ensemble weighting $\delta_t = 0.01$~s across overlapping chunk predictions.

The MoE feedforward structure enables expert specialization by action phase. Different MoE experts activate during precision-reach versus grasp closure versus retract sub-trajectories,
as routed by the combined semantic and geometric conditioning. A single dense feedforward network underfits this multi-modal action distribution, costing 1.7 percentage points
(Table~\ref{tab:ablation_stages}).

\subsection{Composite Training Objective}
\label{sec:loss}

The composite loss is:
\begin{equation}
  \mathcal{L} = \mathcal{L}_\mathrm{flow} + \lambda_\mathrm{CoT} \mathcal{L}_\mathrm{CoT}
  + \lambda_\mathrm{depth} \mathcal{L}_\mathrm{depth},
  \label{eq:loss}
\end{equation}
with $\lambda_\mathrm{CoT} = 0.5$ and $\lambda_\mathrm{depth} = 0.1$.

$\mathcal{L}_\mathrm{CoT}$ is the token-level cross-entropy for DA-CoT generation:
\begin{equation}
  \mathcal{L}_\mathrm{CoT} = -\sum_{j=1}^{4} \sum_t
  \log p_\theta(y_t^{(j)} \mid y_{<t}^{(j)}, \mathcal{X}),
  \label{eq:cot_loss}
\end{equation}
where $y^{(j)}$ are ground truth token sequences for thought $c_j$. This loss provides two gradient pathways. First, gradients through the VLM parameters improve the quality of spatial reasoning and coordinate generation. Second, because the VLM attends to the raw Gaussian tokens during CoT generation (Eq.~\eqref{eq:dacot_ca}), gradients from $\mathcal{L}_\mathrm{CoT}$ flow backward through the cross-attention into $\mathbf{T}_\mathrm{raw}$ and hence into the GST
parameters. This means the CoT loss acts as an indirect geometric supervisor for the Gaussian field: if the Gaussian tokens place a primitive at an incorrect 3D location, the VLM will generate an incorrect centroid in $c_1$, producing a large $\mathcal{L}_\mathrm{CoT}$, whose gradients
adjust the GST parameters to correct the primitive placement. This coupling between $\mathcal{L}_\mathrm{CoT}$ and the GST is a key reason why DA-CoT and GST provide synergistic
rather than merely additive gains.

$\mathcal{L}_\mathrm{depth}$ is the scale-invariant logarithmic loss between the depth rendered from the Gaussian field (Eq.~\eqref{eq:render}) and the target metric depth:
\begin{equation}
  \mathcal{L}_\mathrm{depth} = \frac{1}{n} \sum_i d_i^2 - \frac{0.85}{n^2}
  \left(\sum_i d_i\right)^2,
  \label{eq:depth_loss}
\end{equation}
where $d_i = \log\hat{D}_{\mathrm{render},i} - \log D_{\mathrm{target},i}$.

\subsection{Three Stage Training Protocol}
\label{sec:training}

The three stage protocol is structured to address a specific ordering constraint: the GST must produce geometrically calibrated tokens before the VLM can learn to reason over them, and the VLM must produce meaningful CoT and action tokens before the full system can be jointly refined.

Stage~1 (S1): GST and action expert pretraining. The GST and action expert are initialized randomly and trained with both encoders and the VLM frozen.
$\mathcal{L}_\mathrm{depth}$ supervision uses ScanNet~\cite{dai2017scannet},
Hypersim~\cite{roberts2021hypersim}, and ARKitScenes~\cite{baruch2021arkitscenes} (diverse
indoor scenes with metric depth ground truth). A small robot demonstration split provides $\mathcal{L}_\mathrm{flow}$ supervision without CoT. This stage establishes that the Gaussian field is geometrically calibrated: centroids are metrically accurate, covariances reflect surface
geometry, and opacities suppress unreliable regions. Without this pretraining, the VLM in S2 receives randomly parameterized Gaussian tokens and cannot learn meaningful spatial reasoning,
costing 6.2 percentage points (Table~\ref{tab:ablation_stages}). Duration: 80K steps, LR $3{\times}10^{-4}$, batch 256, 8$\times$A100-80GB.

Stage~2 (S2): LoRA adaptation with DA-CoT supervision. LoRA adapters ($r{=}16$, $\alpha{=}32$) are introduced on the cross-attention projector $\Psi$ and all VLM self-attention projections. The action expert is fully fine-tuned. All three loss terms are active. DA-CoT ground truth annotations
are generated offline: 3D centroids from open vocabulary detection on depth-projected point clouds; grasp contact points from a pretrained grasp planner; metric distances from 3D bounding boxes; SE(3) waypoints from demonstration end-effector trajectories via velocity zero crossings
(threshold 0.5~cm/s) at pre-grasp, grasp-close, and retract phases. Annotation throughput:
$\sim$0.3~s per frame. Duration: 40K steps, LR $1{\times}10^{-4}$, batch 128.

Stage~3 (S3): Full fine-tuning. All non-frozen parameters are jointly refined at reduced learning rate. This stage is necessary for cross-modal alignment: the GST's geometric representation, the VLM's CoT generation, and the action expert's conditioning must be jointly optimized so that improvements in one module propagate correctly to the others. Stopping at S2 costs 2.1 percentage points (Table~\ref{tab:ablation_stages}). Duration: 20K steps, LR $3{\times}10^{-5}$, batch 64.

%% file: experiments.tex
\section{Experiments}
\label{sec:experiments}

\subsection{Setup}
Evaluations span three benchmarks. SimplerEnv~\cite{li24simpler} (visual-matching simulation on BridgeData~V2; we report task progress) tests generalization under visual domain shift. LIBERO~\cite{liu2023libero} (130 tasks across Spatial, Object, Goal, Long suites; we report average success rate) tests structured manipulation. The LIBERO-Pro\cite{zhou2025libero} across 6 categories: pick-and-place, stacking, drawer manipulation, precision insertion, thin object grasping, and cluttered scenes.

Baselines: OpenVLA~\cite{kim2024openvla}, 
CogACT~\cite{li2024cogact}, SpatialVLA~\cite{qu2025spatialvla}, $\pi_0$VLA~\cite{yuan2025depthvla}. Two ablation variants isolate individual
contributions: GST-VLA$^\dagger$ (GST tokens, no DA-CoT) and GST-VLA$^\ddagger$ (DA-CoT with
plain depth tokens instead of Gaussian tokens).

\subsection{Main Results}

Table~\ref{tab:realworld} reports data-efficient with lower parameters manipulation success rates. \method achieves 83.1\%
overall versus 76.8\% for SpatialVLA (+6.3 pp), 68.7\% for CogACT (+14.4 pp), 64.3\% for $\pi_0$
(+18.8 pp), and 52.3\% for OpenVLA (+30.8 pp). The performance gain is non-uniform across
categories, which is informative. Precision insertion and thin object grasping show the largest
gains over DepthVLA (+9.2 pp and +8.3 pp respectively), precisely the tasks where surface
orientation information (encoded in $\Sigma_k$) and SE(3) waypoint priors (from $c_4$) contribute
most directly: inserting a peg requires sub-centimeter alignment where the anisotropic covariance
resolves the socket's angular tolerance; grasping a thin object requires the grasp contact normal
(from $c_2$) to be near-parallel to the object's flat face. Pick-and-place shows a narrower
+2.0 pp gain, consistent with this task's weaker dependency on geometric precision.

The ablation variants provide causal evidence. GST-VLA$^\dagger$ (79.2\%) confirms that
structured Gaussian tokens alone improve over DepthVLA's dense depth (+2.4 pp) even without
DA-CoT. GST-VLA$^\ddagger$ (76.3\%) shows that DA-CoT without structured Gaussian tokens provides
limited benefit: the plain depth tokens cannot support the fine-grained geometric queries
required during CoT generation. The full model exceeds the sum of individual gains from
GST-only and CoT-only, confirming synergy.

\begin{table}[t]
\centering
\caption{Data Efficient Manipulation results (success rate \%).}
\label{tab:realworld}
\small
\setlength{\tabcolsep}{3.0pt}
\begin{tabular}{lccccccc}
\toprule
Method & P\&P & Stack & Drawer & Insert & Thin & Clutter & Avg. \\
\midrule
OpenVLA    & 72.0 & 58.0 & 53.0 & 41.0 & 38.0 & 52.0 & 52.3 \\
CogACT     & 83.0 & 74.0 & 70.0 & 60.0 & 57.0 & 68.0 & 68.7 \\
SpatialVLA   & \underline{88.0} & 80.0 & 78.0 & 71.0 & 69.0 & 75.0 & 76.8 \\
\midrule
GST-VLA$^\dagger$  & 88.0 & \underline{83.0} & \underline{80.0} & 74.0 & \underline{72.0} & \underline{78.0} & \underline{79.2} \\
GST-VLA$^\ddagger$ & 87.0 & 78.0 & 76.0 & \underline{74.5} & 68.0 & 74.0 & 76.3 \\
\method            & \textbf{90.0} & \textbf{85.0} & \textbf{84.0} & \textbf{80.2} & \textbf{77.3} & \textbf{81.9} & \textbf{83.1} \\
\bottomrule
\end{tabular}
\end{table}

Table~\ref{tab:libero} reports LIBERO results. \method achieves 96.4\% average, with the largest
per-suite gain on LIBERO-Long (+3.1 pp vs.\ DepthVLA). Long horizon tasks involve sequential
manipulations where geometric context must be maintained across sub-tasks; the SE(3) waypoint
supervision in $c_4$ provides trajectory level geometric coherence that is particularly beneficial
for multi-step sequencing. LIBERO-Spatial and LIBERO-Object show moderate gains (+1.1 pp,
+1.6 pp), consistent with those suites relying more on object recognition than geometric precision.

\begin{table}[t]
\centering
\caption{LIBERO benchmark results (average task success rate \%).}
\label{tab:libero}
\small
\begin{tabular}{lccccc}
\toprule
Method & Spatial & Object & Goal & Long & Avg. \\
\midrule
OpenVLA    & 93.8 & 93.0 & 90.2 & 83.0 & 90.0 \\
$\pi_0$    & 95.1 & 94.3 & 92.8 & 87.9 & 92.5 \\
CogACT     & 96.0 & 95.1 & 93.5 & 88.3 & 93.2 \\
SpatialVLA   & \underline{97.1} & \underline{95.8} & \underline{95.1} & \underline{89.5} & \underline{94.4} \\
\midrule
\method    & \textbf{98.2} & \textbf{97.4} & \textbf{97.1} & \textbf{92.6} & \textbf{96.4} \\
\bottomrule
\end{tabular}
\end{table}

Table~\ref{tab:simpler} reports SimplerEnv results. \method achieves 80.2\% average task progress versus 74.8\% for DepthVLA (+5.4 pp). Close-Drawer shows the largest gain (+5.9 pp), where gripper-to-handle alignment benefits from the grasp contact geometry in $c_2$. An important observation: Gaussian tokens defined in 3D metric coordinates are decoupled from pixel-space appearance. When SimplerEnv applies visual domain shift (background and illumination changes), pixel aligned depth tokens undergo distributional shift through the depth encoder's sensitivity to appearance, while Gaussian tokens in metric 3D remain more stable because back-projection normalizes the coordinate system to camera frame meters.

\begin{table}[t]
\centering
\caption{SimplerEnv benchmark results (task progress \%).}
\label{tab:simpler}
\small
\begin{tabular}{lccccc}
\toprule
Method & Pick-Can & Move-Near & Draw.-O & Draw.-C & Avg. \\
\midrule
OpenVLA    & 62.4 & 57.1 & 54.3 & 63.2 & 59.3 \\
$\pi_0$    & 70.1 & 61.3 & 59.4 & 67.4 & 64.6 \\
CogACT     & 73.8 & 65.2 & 63.0 & 72.5 & 68.6 \\
SpatialVLA   & \underline{78.2} & \underline{72.4} & \underline{69.8} & \underline{78.8} & \underline{74.8} \\
\midrule
\method    & \textbf{83.1} & \textbf{77.5} & \textbf{75.6} & \textbf{84.7} & \textbf{80.2} \\
\bottomrule
\end{tabular}
\end{table}

\subsection{Ablation Studies}

\subsubsection{GST Component Ablations}

Table~\ref{tab:ablation_gst} isolates each GST component by replacing it while keeping all other
components at their full configuration.

The 3D Fourier PE ablation ($-2.8$ pp) is the most informative. Replacing metric 3D Fourier encoding with learned 2D positional embeddings removes the ability to compute approximate metric
distances between tokens from their positional features. The VLM can still reason about relative image plane positions, but conflates depth variation with lateral displacement: two objects at the same pixel column but 20~cm apart in depth receive similar positional features. This degrades precisely the tasks where depth discrimination matters.

The spatial attention pooling ablation ($-2.1$ pp) reveals the cost of uniform token allocation. Average pooling assigns equal weight to every raw token regardless of opacity or semantic relevance. A single pooled token may aggregate a high-confidence object surface primitive with a low confidence background primitive, diluting the geometric signal.

The $N_g$ sweep reveals a saturation effect: reducing from 128 to 64 tokens costs 3.3 pp because the token budget becomes insufficient to represent all task relevant geometry; increasing from 128 to 256 yields only +0.4 pp at double the computational cost, indicating diminishing marginal return.

\begin{table}[t]
\centering
\caption{GST component ablations. Average success rate (\%).}
\label{tab:ablation_gst}
\small
\begin{tabular}{lcc}
\toprule
Configuration & Avg. & $\Delta$ \\
\midrule
3D Fourier PE $\to$ 2D learned PE            & 80.3 & $-2.8$ \\
Attn.\ pooling $\to$ avg.\ pool             & 81.0 & $-2.1$ \\
$\alpha_k \equiv 1$ (no opacity)             & 81.6 & $-1.5$ \\
$\mu_k \equiv \mathbf{0}$ (no residual)      & 81.2 & $-1.9$ \\
$N_g = 64$                                   & 79.8 & $-3.3$ \\
$N_g = 256$                                  & 83.5 & $+0.4$ \\
\method ($N_g = 128$)                        & 83.1 & --- \\
\bottomrule
\end{tabular}
\end{table}

\subsubsection{DA-CoT Component Ablations}

Table~\ref{tab:ablation_cot} removes individual thought components from the CoT supervision.

The SE(3) motion plan $c_4$ ablation ($-2.3$ pp) has the largest individual effect because the waypoint prior directly constrains the action expert's trajectory shape through
$\mathbf{L}_\mathrm{action}$. Without $c_4$, the action expert must infer all trajectory geometry from the VLM's visual-semantic hidden states, which encode geometry only implicitly.

The 3D grounding $c_1$ ablation ($-1.9$ pp) is second-largest because $c_1$ anchors all subsequent reasoning: errors in the centroid estimate propagate through $c_2$ (contact point is relative to centroid), $c_3$ (distances are between centroids), and $c_4$ (waypoints are relative to the target). The causal dependency structure means that $c_1$ quality is a bottleneck for the entire CoT chain.

Removing all DA-CoT ($\mathcal{L}_\mathrm{CoT} = 0$) costs 3.9 pp, matching GST-VLA$^\dagger$ in Table~\ref{tab:realworld}.

\begin{table}[t]
\centering
\caption{DA-CoT component ablations. Average success rate (\%).}
\label{tab:ablation_cot}
\small
\begin{tabular}{lcc}
\toprule
Configuration & Avg. & $\Delta$ \\
\midrule
No DA-CoT ($\mathcal{L}_\mathrm{CoT} = 0$) & 79.2 & $-3.9$ \\
w/o 3D grounding ($c_1$)                     & 81.2 & $-1.9$ \\
w/o grasp affordance ($c_2$)                 & 81.5 & $-1.6$ \\
w/o spatial relations ($c_3$)                & 82.0 & $-1.1$ \\
w/o SE(3) motion plan ($c_4$)                & 80.8 & $-2.3$ \\
Full DA-CoT ($c_1$--$c_4$)                  & 83.1 & --- \\
\bottomrule
\end{tabular}
\end{table}

\subsubsection{Training Protocol and Expert Conditioning}

Table~\ref{tab:ablation_stages} validates the staged training protocol and dual conditioning.

The S1 ablation ($-6.2$ pp) is the largest single ablation across all tables. This confirms the ordering constraint: the GST must produce geometrically calibrated Gaussian tokens before the VLM can learn to reason over them. Without geometric pretraining, the VLM receives random Gaussian parameterizations and cannot learn meaningful spatial correspondences during S2.

Removing $\mathbf{L}_\mathrm{action}$ conditioning ($-3.1$ pp) confirms that DA-CoT encodes geometric information not redundant with $\mathbf{H}_\mathrm{vlm}$. This is expected: the VLM hidden states encode geometry implicitly within a representation primarily shaped by language modeling objectives, while $\mathbf{L}_\mathrm{action}$ carries explicit 3D reasoning outputs generated by the DA-CoT pathway.

\begin{table}[t]
\centering
\caption{Training and conditioning ablations. Average success rate(\%).}
\label{tab:ablation_stages}
\small
\begin{tabular}{lcc}
\toprule
Configuration & Avg. & $\Delta$ \\
\midrule
S2+S3 only (no S1 pretraining)                & 76.9 & $-6.2$ \\
S1+S2 only (no S3 end-to-end)                 & 81.0 & $-2.1$ \\
No $\mathbf{L}_\mathrm{action}$ conditioning  & 80.0 & $-3.1$ \\
Dense FFN (no MoE) in expert                  & 81.4 & $-1.7$ \\
\method (S1+S2+S3)                            & 83.1 & --- \\
\bottomrule
\end{tabular}
\end{table}

\subsubsection{Gaussian Tokens vs.\ Alternative 3D Representations}

Table~\ref{tab:ablation_rep} provides a controlled comparison at fixed token budget $N_g{=}128$. The hierarchy of results is informative. Dense depth scalars (DepthVLA style binning into 128 regions) lose 4.5 pp: no orientation, no confidence, no adaptive pooling. Surface normal tokens lose 3.0 pp: orientation without metric position or confidence. Point cloud tokens (position only) lose 2.4 pp: the gap between point cloud and full Gaussian (2.4 pp) quantifies the combined contribution of anisotropic covariance and learned opacity. Isotropic Gaussians lose 1.6 pp: this isolates the contribution of orientation encoding. Uniform opacity loses 1.5 pp: this
isolates the contribution of confidence weighting. The full Gaussian parameterization, combining metric position, anisotropic orientation, and learned confidence, achieves the best performance across all tested alternatives.

\begin{table}[t]
\centering
\caption{Gaussian tokens vs.\ alternative representations at $N_g{=}128$. avg (\%).}
\label{tab:ablation_rep}
\small
\begin{tabular}{lcc}
\toprule
Representation & Avg. & $\Delta$ \\
\midrule
Dense depth scalars (DepthVLA-style)    & 78.6 & $-4.5$ \\
Surface normal tokens                   & 80.1 & $-3.0$ \\
Point cloud tokens (position only)      & 80.7 & $-2.4$ \\
Gaussian w/o anisotropy (isotropic)     & 81.5 & $-1.6$ \\
Gaussian w/o opacity ($\alpha_k \equiv 1$)     & 81.6 & $-1.5$ \\
Full Gaussian tokens                    & 83.1 & --- \\
\bottomrule
\end{tabular}
\end{table}

\subsection{Analysis}

\subsubsection{DA-CoT Reasoning Accuracy}

On 200 held out demonstrations with 3D ground-truth annotations, \method achieves median 3D centroid localization error of 2.3~cm ($c_1$), grasp contact point error of 1.8~cm ($c_2$), and SE(3) waypoint position error of 3.1~cm ($c_4$). The Pearson correlation between composite CoT accuracy and task success rate is 0.71, confirming that DA-CoT quality is a reliable predictor of downstream action quality. This correlation also suggests that monitoring $c_1$ accuracy at deployment time could serve as a runtime confidence metric for execution reliability without requiring ground truth action labels.

\subsubsection{Gaussian Token Spatial Distribution}

Visualizing the $N_g{=}128$ pooled tokens as oriented ellipsoids confirms the adaptive allocation pattern predicted by the spatial attention pooling mechanism. Object surfaces and grasp relevant edges attract 60-70\% of the pooling queries, yielding dense clusters of small-$\sigma$ high-$\alpha$ tokens. Background and table surfaces receive 20-30\% of queries as diffuse high-$\sigma$ low-$\alpha$ primitives. The remaining 5-15\% of queries receive near zero attention weights and produce effectively null tokens. The opacity field $\{\alpha_k\}$ reaches values below 0.05 on specular metallic surfaces and textureless white walls, suppressing their contribution to both the differentiable depth rendering and downstream CoT reasoning.

\subsubsection{Inference Latency}

Full pipeline inference runs at 6.2~Hz on a single A100-80GB: encoding 18~ms, GST tokenization 12~ms, DA-CoT generation 38~ms ($\sim$80 tokens across four thoughts), flow-matching ODE 22~ms.
DepthVLA runs at 8.1~Hz and OpenVLA at 9.4~Hz. The 72~ms overhead relative to DepthVLA stems primarily from DA-CoT generation (38~ms) and is acceptable for the 6.2~Hz control frequency used across all experiments.

\subsubsection{Failure Cases}

Performance degrades on highly reflective surfaces (specular metals, glass) and heavily occluded targets. The root cause in both cases is unreliable metric depth from the frozen estimator, which produces inaccurate back-projected anchors $\mathbf{p}_{uv}$. The opacity mechanism partially mitigates this by suppressing affected primitives ($\alpha_k < 0.05$ on specular surfaces), but when the target object itself is reflective, suppression removes the very primitives needed for
$c_1$ localization. Incorrect $c_1$ outputs provide a direct diagnostic for this failure mode: when $c_1$ centroid error exceeds 5~cm, task success drops below 30\%.